%% file: main.tex
\documentclass[sigconf]{acmart}

\AtBeginDocument{%
  }

\setcopyright{acmlicensed}
\copyrightyear{2026}
\acmYear{2026}
\setcopyright{cc}
\setcctype{by}
\acmConference[MM '26]{Proceedings of the 34th ACM International Conference on Multimedia}{November 10--14, 2026}{Rio de Janeiro, Brazil}
\acmBooktitle{Proceedings of the 34th ACM International Conference on Multimedia (MM '26), November 10--14, 2026, Rio de Janeiro, Brazil}
\acmDOI{10.1145/3767308.3836065}
\acmISBN{979-8-4007-2213-4/2026/11}

\acmSubmissionID{6420}

\usepackage{booktabs}
\usepackage{multirow}
\usepackage{colortbl}
\usepackage{graphicx}

\begin{document}

\title{AirForesight: Current-to-Future Spatial Map Imagination with Cross-Space Planning Consistency for UAV-VLN}

\author{Yutong Liu}
\authornote{Equal contribution.}
\affiliation{%
  \institution{Harbin Institute of Technology, Shenzhen}
  \city{Shenzhen}
  \country{China}
}
\email{yutongliu0112@gmail.com}

\author{Xiaojie Li}
\authornotemark[1]
\authornote{Corresponding authors.}
\affiliation{%
  \institution{Harbin Institute of Technology, Shenzhen}
  \city{Shenzhen}
  \country{China}
}
\email{xiaojieli0903@gmail.com}

\author{Mingzhu Xu}
\affiliation{%
  \institution{Shandong University}
  \city{Jinan}
  \country{China}
}
\email{xumingzhu@sdu.edu.cn}


\author{Jianlong Wu}
\authornotemark[2]
\affiliation{%
  \institution{Harbin Institute of Technology, Shenzhen}
  \institution{Shenzhen Loop Area Institute}
  \city{Shenzhen}
  \country{China}
}
\email{wujianlong@hit.edu.cn}

\renewcommand{\shortauthors}{Yutong Liu, Xiaojie Li, Mingzhu Xu, and Jianlong Wu}

\begin{abstract}
Unmanned Aerial Vehicle Vision-Language Navigation (UAV-VLN) requires agents to follow language instructions, infer spatial structure from sparse multi-view observations, and execute feasible 3D motion in complex outdoor environments. Despite recent progress with large language models, most existing methods still map vision-language inputs directly to actions, providing limited explicit scene grounding and future-aware spatial reasoning.
We propose \textbf{AirForesight}, a current-to-future spatial map imagination framework for UAV-VLN. AirForesight first learns a structured current-map representation from multi-view observations. This representation is jointly supervised by current-map reconstruction and future-trajectory prediction, encouraging it to encode both present scene structure and future motion intent. Under structured causal attention, the current spatial knowledge is propagated to future-map reasoning, and the resulting current and future representations are aggregated to predict the next 3D waypoint. To make spatial imagination more relevant to navigation, we introduce a \textbf{cross-space planning consistency loss} that encourages directional agreement between the predicted map-space trajectory and the expert action direction derived from the ground-truth waypoint displacement.
Experiments on OpenUAV and AerialVLN-S, together with extensive ablations, demonstrate strong performance and support the effectiveness and stability of the proposed framework.
\end{abstract}

\ccsdesc[500]{Computing methodologies~Cognitive robotics}
\ccsdesc[300]{Computing methodologies~Robotic planning}
\ccsdesc[100]{Computing methodologies~Scene understanding}

\keywords{Vision-Language Navigation, Unmanned Aerial Vehicle, Large Language Model, Spatial Map Imagination}


\maketitle

\newcommand{\modelName}{AirForesight}
\newcommand{\curmap}{CSMM}
\newcommand{\futuremap}{FSI}
\newcommand{\loss}{CPCL}

\section{Introduction}

\begin{figure}[!t]
    \centering
    \includegraphics[width=0.5\textwidth]{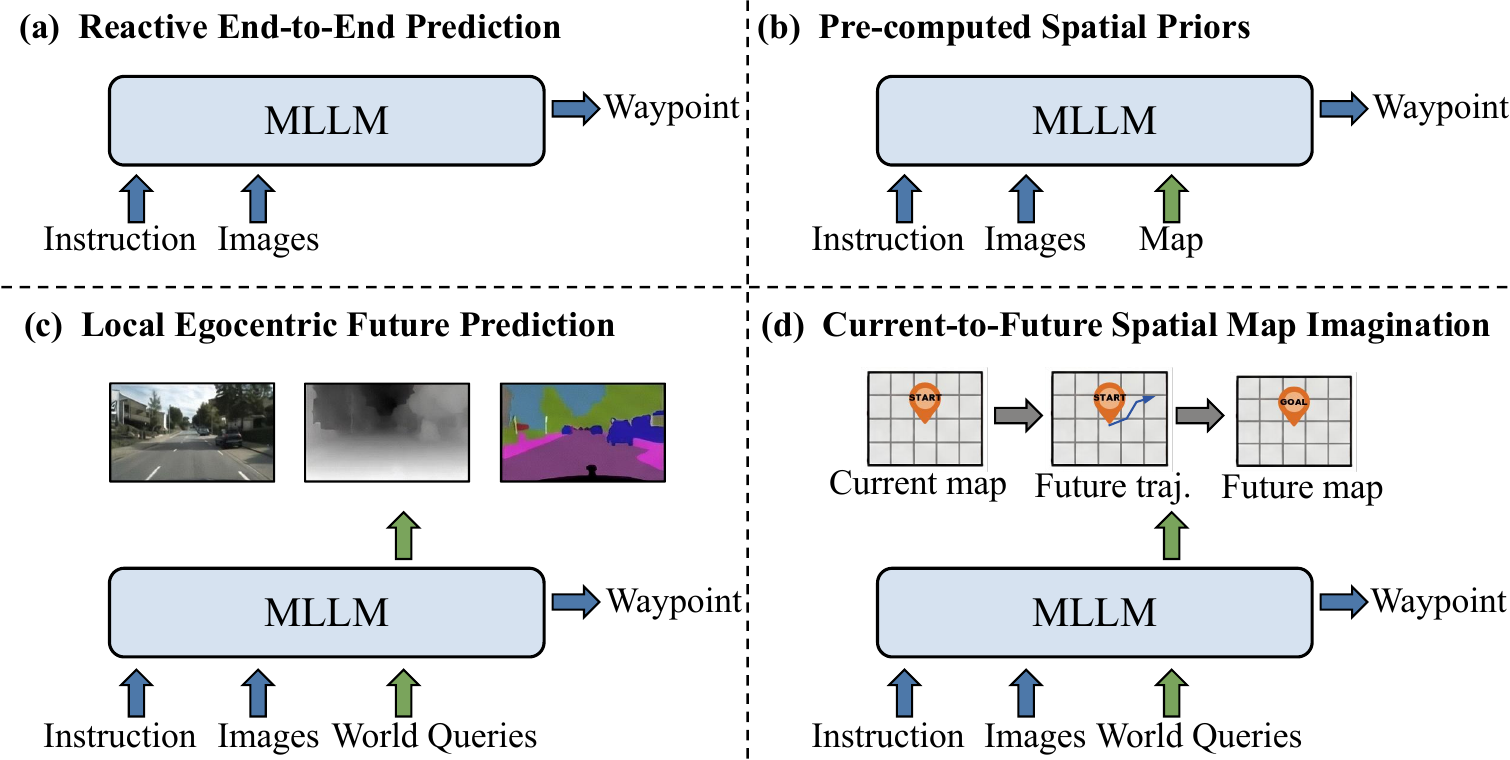}
    \caption{Comparison of existing paradigms for UAV-VLN.}
    \label{fig:existing_pipeline}
    
    \vspace{-4mm}
\end{figure}

Unmanned Aerial Vehicle (UAV) Vision-Language Navigation (VLN)~\cite{aerialvln,traveluav,openfly,citynav} aims to enable an aerial agent to navigate complex environments following natural-language instructions. This task requires the coordination of visual perception, instruction grounding, trajectory planning, and action execution~\cite{shi2026intelligent}, with broad potential in applications such as search and rescue, infrastructure inspection, and autonomous delivery.

Earlier UAV-VLN methods mainly rely on recurrent or cross-modal attention policies~\cite{aerialvln}, while recent approaches employ large language models (LLMs) or multimodal large language models as navigation backbones to improve instruction understanding and high-level decision-making~\cite{traveluav,openfly}. Nevertheless, many existing methods still predict waypoints directly from language instructions and multi-view observations. This design presents three limitations in complex, long-horizon aerial navigation:

\textbf{(1) Insufficient current-scene grounding.} As illustrated in Figure~\ref{fig:existing_pipeline}(a), recent methods commonly predict waypoints from discrete multi-view observations without organizing them into a structured spatial state~\cite{traveluav,openfly}. Consequently, agents may over-rely on short-term visual cues and struggle to maintain a coherent semantic--geometric understanding of their surroundings. Some approaches instead construct geometric or semantic maps as spatial priors~\cite{mapnav,vlmaps,STMR}, as shown in Figure~\ref{fig:existing_pipeline}(b). However, constructing and encoding these maps requires additional preprocessing and increases navigation latency when performed online. \textbf{(2) Limited future spatial foresight.} Reliable UAV-VLN requires anticipating both future motion and the spatial states that may be encountered. Existing methods forecast RGB images, depth maps, semantic observations, or latent states~\cite{cot-vla,dreamvla,pathdreamer,vista,viewpoint-1,navcot}, as illustrated in Figure~\ref{fig:existing_pipeline}(c), but these predictions are generally confined to local egocentric views and provide limited modeling of future motion and map-level environmental evolution within a shared spatial representation. \textbf{(3) Weak connection between spatial imagination and action execution.} Even when intermediate spatial predictions are introduced, they are often optimized as auxiliary objectives and remain only weakly connected to downstream action generation~\cite{predtraj-1,vlfm}. Without an explicit consistency constraint, map-space trajectory reasoning may provide insufficiently aligned guidance for executable UAV navigation.

To address these limitations, we propose \textbf{AirForesight}, a \textbf{current-to-future spatial map imagination} framework for UAV-VLN. AirForesight first organizes multi-view observations into a latent current-map representation that captures the semantic and geometric structure of the surrounding environment. This representation is jointly supervised to reconstruct the current map and predict a future trajectory in map space, encouraging the model to encode both present scene structure and future motion intent. A structured causal attention mechanism then propagates the current spatial knowledge to future-map reasoning, allowing the model to infer how the environment may evolve after motion. Finally, the current and anticipated spatial representations are aggregated to predict the next 3D waypoint. In this way, AirForesight grounds navigation decisions in current scene structure while enabling future-aware spatial reasoning, with spatial maps used only for training supervision rather than online dense-map construction.

To make the predicted map-space trajectory more relevant to executable navigation, we further introduce a \textbf{cross-space planning consistency loss}. The loss encourages directional agreement between the predicted map-space trajectory and the expert action direction derived from the ground-truth waypoint displacement. It thereby bridges map-space reasoning and action-space execution, improving the consistency between spatial planning and navigation decisions.

Our main contributions are summarized as follows:
\vspace{-4mm}
\begin{itemize}
    \item We propose AirForesight, a current-to-future spatial map imagination framework for UAV-VLN that learns latent  representations of the current spatial map and future trajectory, progressively anticipates the future spatial map representations, and integrates these representations for waypoint prediction.

    \item We introduce a cross-space planning consistency loss that encourages directional agreement between the predicted map-space trajectory and the expert action direction, linking spatial imagination with downstream waypoint prediction.

    \item On the OpenUAV Test Seen split, AirForesight improves SR from 30.47\% to 35.83\% and SPL from 25.37\% to 30.22\%, while reducing NE from 68.44\,m to 56.99\,m over TravelUAV. Additional results on AerialVLN-S, multi-seed experiments, and detailed ablations further support the effectiveness and stability of the framework.
\end{itemize}

\section{Related Works}
\subsection{Vision-Language Navigation}

Vision-Language Navigation (VLN) has been extensively studied in ground-based embodied environments through diverse benchmarks~\cite{touchdown,beyondgraph,staypath,room,reverie} and advances in memory modeling~\cite{history1,history2,history3,history4}, data augmentation~\cite{dataaug-1,dataaug-2,dataaug-3}, and policy learning~\cite{action-reinforced,action-target}. Beyond direct action prediction, many approaches introduce structured spatial representations, including topological maps~\cite{topo-1,topo-2}, semantic maps~\cite{vlmaps,mapnav,vlfm}, point clouds~\cite{openset3d,openscene,clip2scene}, and hybrid representations~\cite{bevbert}. Although effective, these approaches often require explicit spatial reconstruction and continual updates, introducing additional computation and latency.

UAV-VLN extends this task to aerial environments with larger viewpoint changes, longer navigation horizons, and continuous 3D action spaces. AerialVLN~\cite{aerialvln} and CityNav~\cite{citynav} introduce representative aerial navigation benchmarks, where earlier methods mainly rely on recurrent networks and cross-modal attention for action prediction. More recent approaches, such as TravelUAV~\cite{traveluav} and OpenFly~\cite{openfly}, employ multimodal large language models to improve instruction understanding and high-level navigation planning. However, these methods still largely predict waypoints directly from language instructions and multi-view observations, without first forming a structured representation of the surrounding semantic and geometric layout or its future evolution. Some approaches introduce top-down semantic representations or bird's-eye-view priors~\cite{STMR,grid-based-view-selection}, but they typically rely on pre-computed or densely constructed spatial inputs. In contrast, AirForesight learns compact latent spatial representations for current-scene understanding, future-state reasoning, and waypoint prediction without online dense reconstruction.

\begin{figure*}[!th]
    \centering
    \includegraphics[width=\textwidth]{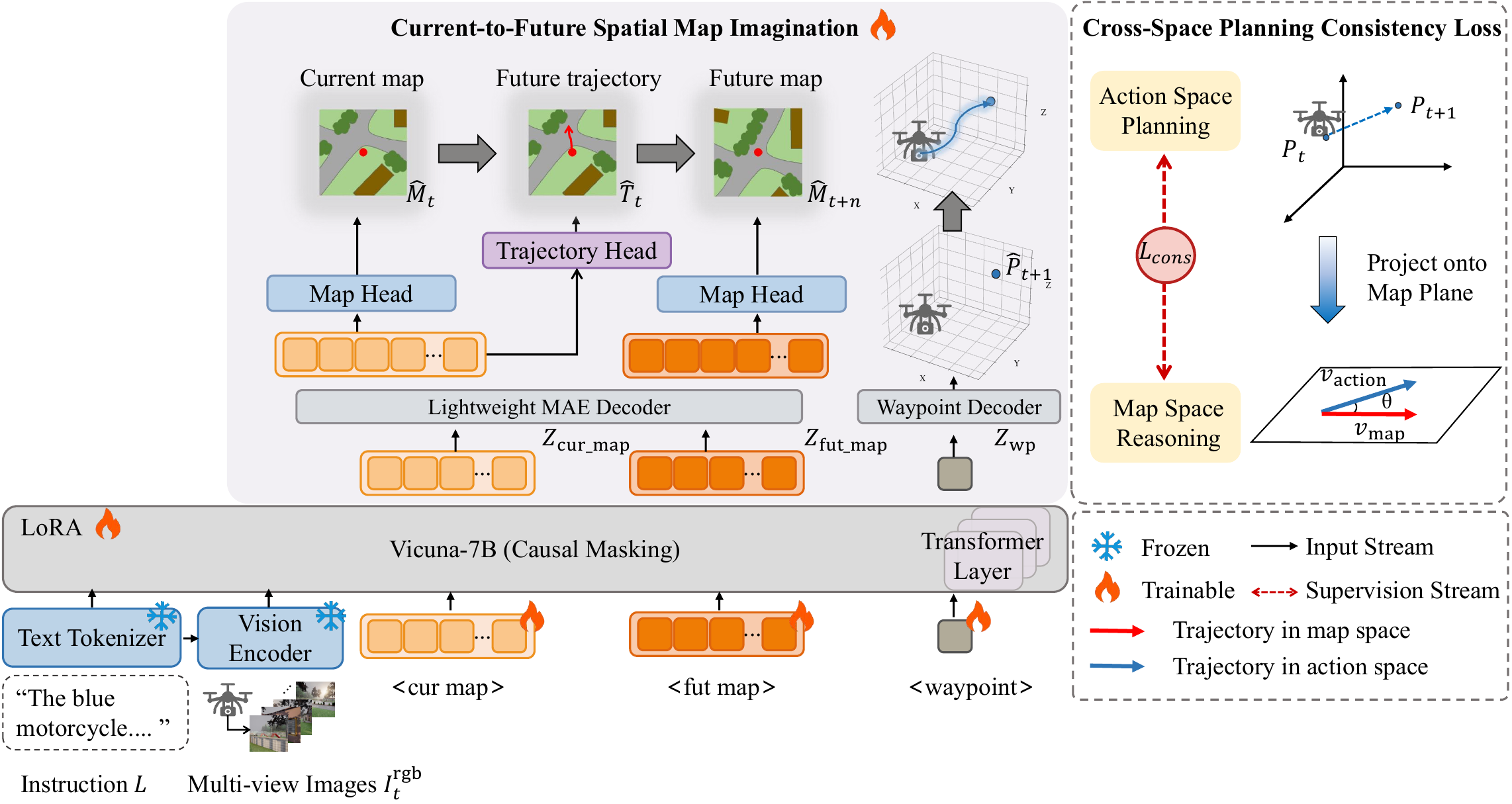}
    \caption{Overall architecture of AirForesight. The model jointly processes multimodal inputs with learnable current-map, future-map, and waypoint tokens. The current-map representation is supervised by current-map reconstruction and future-trajectory prediction, while structured causal attention propagates current spatial knowledge to the future-map representation, which is optimized by future-map reconstruction. The waypoint representation integrates current and future spatial cues to predict the next 3D waypoint. A cross-space planning consistency loss further enforces directional alignment between the predicted map-space trajectory and the expert action direction.}
    \label{fig:overall}
\end{figure*}

\subsection{Visual Imagination}

Visual imagination enables embodied agents to anticipate future states before acting. Existing approaches can be broadly grouped according to how future information is represented.

\textbf{Local observation forecasting} predicts future egocentric RGB frames, depth maps, or semantic segmentations~\cite{cot-vla,dreamvla,pathdreamer,vista,viewpoint-1,navcot}. These methods provide short-term foresight but remain confined to local first-person views and do not explicitly capture route-level spatial evolution.

\textbf{Implicit future representation modeling} predicts future states in compact latent spaces~\cite{navforesee,navq,navmorph}. Although efficient, these representations are often weakly grounded in explicit geometry, making their connection to feasible 3D navigation less direct.

\textbf{Explicit spatial imagination} predicts or completes structured spatial layouts to support planning~\cite{sea,mapdream,predtraj-1}. These methods provide clearer spatial context but mainly focus on current-state reconstruction or spatial completion, rather than jointly modeling present scene structure, future motion intent, and future environmental states within a causally ordered process. Their intermediate spatial predictions are also often only weakly connected to downstream action generation.

AirForesight instead jointly learns representations of the current spatial structure, future trajectory, and future spatial state under structured causal attention. It further introduces a cross-space planning consistency loss that encourages agreement between the predicted map-space trajectory direction and the expert action direction, making spatial imagination more relevant to waypoint prediction.
\vspace{-0.5mm}

\section{Method}

\subsection{Problem Formulation}

UAV-VLN requires an aerial agent to follow a natural-language instruction and navigate toward a target location in a 3D environment. At each time step, the UAV receives multi-view RGB observations
$
I_t^{\mathrm{rgb}}=\{I_{t,v}^{\mathrm{rgb}}\}_{v\in\mathcal{V}},
$
where
$
\mathcal{V}=\{\text{front},\text{rear},\text{left},\text{right},\text{down}\},
$
together with a language instruction $L$.
Let
$
P_t=[x_t,y_t,z_t]
$
denote the UAV position at time step $t$, with $P_0$ representing its initial position. The goal is to learn a navigation policy $\pi$ that predicts the next 3D waypoint $\hat{P}_{t+1}$ from the multimodal inputs:
\begin{equation}
    \pi:(I_t^{\mathrm{rgb}},L)\rightarrow\hat{P}_{t+1}.
\end{equation}
The predicted waypoint is executed in a closed-loop manner, with the UAV receiving new observations and predicting subsequent waypoints iteratively. The process terminates when the stopping condition is met or the maximum navigation horizon is reached. An episode is successful if the final UAV position falls within a predefined distance threshold of the target.

\subsection{Overall Architecture}

As illustrated in Figure~\ref{fig:overall}, AirForesight performs current-to-future spatial map imagination before predicting the next 3D waypoint. The framework consists of multimodal input encoding, current-to-future spatial representation learning, waypoint prediction, and cross-space planning consistency.

The language instruction $L$ and multi-view RGB observations $I_t^{\mathrm{rgb}}$ are first encoded into a unified multimodal token sequence. Three groups of learnable query tokens---current-map, future-map, and waypoint tokens---are appended to this sequence and processed by the Vicuna-7B backbone~\cite{vicuna} under a causal attention mask.

The current-map tokens attend to the multimodal inputs and produce a latent current-map representation. A lightweight MAE decoder and two task-specific heads decode this representation into the current map $\hat{M}_t$ and an $n$-step future trajectory $\hat{T}_t$ in map space. Joint supervision from these two objectives encourages the current-map representation to encode both present scene structure and future motion intent.
The future-map tokens attend to both the multimodal inputs and the preceding current-map tokens, producing a latent future-map representation that is decoded into the future map $\hat{M}_{t+n}$. 
The waypoint token is placed at the end of this causal sequence and aggregates the multimodal, current-map, and future-map representations to predict the next 3D waypoint $\hat{P}_{t+1}$, which is subsequently refined into a feasible local path. The detailed formulation is presented in Sec.~\ref{sec:cf_map_imagination}.

To connect map-space reasoning with action-space supervision, we introduce a \textbf{cross-space planning consistency loss}. It encourages directional agreement between the predicted map-space trajectory and the expert action direction derived from the ground-truth waypoint displacement, making the learned spatial representation more relevant to downstream waypoint prediction. Its formulation is presented in Sec.~\ref{sec:cross_space_consistency}.

The entire framework is trained with a unified objective that jointly optimizes waypoint prediction, current-map reconstruction, future-trajectory prediction, future-map prediction, and cross-space planning consistency, as described in Sec.~\ref{sec:overall_optimization}.

\subsection{Current-to-Future Spatial Map Imagination}
\label{sec:cf_map_imagination}

Existing UAV-VLN methods often predict actions directly from multi-view observations without explicitly representing either the current spatial layout or its future evolution. AirForesight instead learns a causally ordered sequence of three structured intermediate states: a current map, a future trajectory in map space, and a future map. The current-map representation is jointly supervised by current-map reconstruction and future-trajectory prediction, while structured causal attention propagates this representation to future-map reasoning and waypoint prediction.

\subsubsection{Current Spatial Map Modeling}

Let $\langle\text{cur\_map}\rangle$ denote a set of learnable current-map tokens. These tokens attend to the language instruction $L$ and multi-view RGB observations $I_t^{\mathrm{rgb}}$ through the LLM backbone $\mathcal{M}$, producing a latent current-map representation:
\begin{equation}
Z_{\mathrm{cur\_map}}
=
\mathcal{M}
\left(
L,I_t^{\mathrm{rgb}}
\mid
\langle\text{cur\_map}\rangle
\right).
\end{equation}
A lightweight MAE decoder~\cite{vae} transforms
$Z_{\mathrm{cur\_map}}$ into dense spatial features $F_{\mathrm{cur}}$, which are projected by a linear classification head to produce the current semantic map $\hat{M}_t$.

We construct the supervision maps using an automatic offline annotation pipeline. An LLM extracts task-relevant semantic categories from the instruction $L$. These categories are used to prompt GroundingDINO~\cite{groundingdino} and MobileSAM~\cite{mobilesam} for object grounding and segmentation in the multi-view RGB observations. Using the corresponding depth observations $I_t^{\mathrm{depth}}$ and camera parameters, the resulting semantic masks are unprojected into a 3D point cloud, transformed into the current UAV-local coordinate frame, and orthogonally projected onto a fixed-scale 2D grid. If multiple semantic points occupy the same grid cell, the label of the highest point is retained. Semantically overlapping categories are then consolidated into $C$ superclasses, producing the pseudo-label map
\begin{equation}
M_t\in\{1,\ldots,C\}^{H\times W},
\end{equation}
where $H$ and $W$ denote the grid height and width. The annotation models are used only to generate supervision offline and are not required during model training or inference.

The current-map prediction is optimized using weighted cross-entropy:
\begin{equation}
\mathcal{L}_{\mathrm{map}}^{\mathrm{cur}}
=
-\frac{1}{HW}
\sum_{i=1}^{H}
\sum_{j=1}^{W}
\sum_{c=1}^{C}
w_{1,c}
\mathbf{1}\!\left(M_t^{(i,j)}=c\right)
\log \hat{p}_{t,c}^{(i,j)},
\end{equation}
where $\hat{p}_{t,c}^{(i,j)}$ is the predicted probability that grid cell $(i,j)$ belongs to class $c$, $\mathbf{1}(\cdot)$ is the indicator function, and $w_{1,c}$ reweights class $c$ to alleviate class imbalance.

\subsubsection{Future Trajectory Imagination}

The current spatial features $F_{\mathrm{cur}}$ are also fed into a binary trajectory head to predict an $n$-step future trajectory mask $\hat{T}_t$ in the same map space. The corresponding supervision mask
\begin{equation}
T_t\in\{0,1\}^{H\times W}
\end{equation}
is obtained by projecting the future physical waypoints
$\{P_{t+k}\}_{k=1}^{n}$ onto the current UAV-local map plane.

The trajectory prediction is optimized by weighted binary cross-entropy:
\begin{equation}
\begin{aligned}
\mathcal{L}_{\mathrm{traj}}
=
-\frac{1}{HW}
\sum_{i=1}^{H}
\sum_{j=1}^{W}
\Big[
&w_{2,1}T_t^{(i,j)}
\log \hat{q}_t^{(i,j)}
\\
+{}&
w_{2,0}\left(1-T_t^{(i,j)}\right)
\log\left(1-\hat{q}_t^{(i,j)}\right)
\Big],
\end{aligned}
\end{equation}
where $\hat{q}_t^{(i,j)}$ is the predicted trajectory probability at grid cell $(i,j)$, and $w_{2,1}$ and $w_{2,0}$ balance the sparse trajectory and background classes, respectively.

Because the current-map and future-trajectory heads share $F_{\mathrm{cur}}$, trajectory supervision encourages the current-map representation to encode both present scene structure and future motion intent. The decoded trajectory mask $\hat{T}_t$ is not explicitly re-encoded as an input to the future-map or waypoint branches; instead, its planning information is conveyed through the shared current-map representation.

\subsubsection{Future Map Imagination}

To anticipate the spatial state at step $t+n$, we introduce learnable future-map tokens $\langle\text{fut\_map}\rangle$. Under structured causal attention, these tokens attend to the multimodal inputs and the preceding current-map tokens:
\begin{equation}
Z_{\mathrm{fut\_map}}
=
\mathcal{M}
\left(
L,I_t^{\mathrm{rgb}},
\langle\text{cur\_map}\rangle
\mid
\langle\text{fut\_map}\rangle
\right).
\end{equation}
For clarity, this equation denotes the output of the future-map tokens within the same causally masked forward pass, rather than a separate invocation of the backbone.

The future-map representation is decoded by the MAE decoder and map classifier shared with the current-map branch, producing $\hat{M}_{t+n}$. Its supervision map $M_{t+n}$ is generated using the same offline annotation pipeline as $M_t$. Both maps are represented in the current UAV-local coordinate frame, allowing current and future spatial states to be compared in a shared coordinate system.

The future-map prediction is optimized using
\begin{equation}
\mathcal{L}_{\mathrm{map}}^{\mathrm{fut}}
=
-\frac{1}{HW}
\sum_{i=1}^{H}
\sum_{j=1}^{W}
\sum_{c=1}^{C}
w_{1,c}
\mathbf{1}\!\left(M_{t+n}^{(i,j)}=c\right)
\log \hat{p}_{t+n,c}^{(i,j)}.
\end{equation}

\subsubsection{Waypoint Prediction}

A learnable waypoint token $\langle\text{waypoint}\rangle$ is placed after the current-map and future-map tokens in the causal sequence. It therefore aggregates the multimodal context together with the preceding current- and future-map representations:
\begin{equation}
Z_{\mathrm{wp}}
=
\mathcal{M}
\left(
L,I_t^{\mathrm{rgb}},
\langle\text{cur\_map}\rangle,
\langle\text{fut\_map}\rangle
\mid
\langle\text{waypoint}\rangle
\right).
\end{equation}
Accordingly, future-trajectory supervision influences waypoint prediction through the shared current-map representation, rather than through direct use of the decoded trajectory mask.

An MLP decoder maps $Z_{\mathrm{wp}}$ to the normalized direction
$\hat{\mathbf{u}}_{t+1}\in\mathbb{R}^{3}$ and distance
$\hat{d}_{t+1}\in\mathbb{R}$ of the next waypoint. The navigation loss is
\begin{equation}
\mathcal{L}_{\mathrm{nav}}
=
\left|
\hat{d}_{t+1}-d_{t+1}
\right|
+
1-
\cos
\left(
\hat{\mathbf{u}}_{t+1},
\mathbf{u}_{t+1}
\right),
\end{equation}
where $d_{t+1}$ and $\mathbf{u}_{t+1}$ are the ground-truth waypoint distance and normalized direction, respectively. Given the predicted waypoint, the lightweight path decoder from TravelUAV~\cite{traveluav} generates a fine-grained local route for execution.

\begin{table*}[!t]
\centering
\caption{Comprehensive results on the OpenUAV benchmark under L1-level assistance. ``Seen'' denotes Test Seen, while ``UM'' and ``UO'' denote Test Unseen Map and Test Unseen Object, respectively.}
\label{tab:final_results}

\renewcommand{\arraystretch}{0.95}
\setlength{\tabcolsep}{4.5pt}

\resizebox{\textwidth}{!}{%
\begin{tabular}{lc|cccc|cccc|cccc}
\toprule
\multirow{2}{*}{Method}
& \multirow{2}{*}{Test Set}
& \multicolumn{4}{c|}{Full}
& \multicolumn{4}{c|}{Easy}
& \multicolumn{4}{c}{Hard} \\
\cmidrule(lr){3-6}
\cmidrule(lr){7-10}
\cmidrule(lr){11-14}
& 
& NE$\downarrow$ & SR$\uparrow$ & OSR$\uparrow$ & SPL$\uparrow$
& NE$\downarrow$ & SR$\uparrow$ & OSR$\uparrow$ & SPL$\uparrow$
& NE$\downarrow$ & SR$\uparrow$ & OSR$\uparrow$ & SPL$\uparrow$ \\
\midrule

Random & Seen
& 222.20 & 0.14 & 0.21 & 0.07
& 142.07 & 0.26 & 0.39 & 0.13
& 320.12 & 0.00 & 0.00 & 0.00 \\

Fixed Action & Seen
& 188.61 & 2.27 & 8.16 & 1.40
& 121.36 & 3.48 & 11.48 & 2.14
& 270.69 & 0.79 & 4.09 & 0.49 \\

CMA~\cite{aerialvln} & Seen
& 135.73 & 8.37 & 18.72 & 7.90
& 84.89 & 11.48 & 24.52 & 10.68
& 197.77 & 4.57 & 11.65 & 4.51 \\

NavFoM~\cite{navfom} & Seen
& 93.05 & 29.17 & 49.24 & 25.03
& 58.98 & 32.91 & 53.16 & 27.87
& 143.83 & 23.58 & 43.40 & 20.80 \\

TravelUAV~\cite{traveluav} & Seen
& 68.44 & 30.47 & 61.64 & 25.37
& 47.53 & 32.40 & 63.35 & 25.98
& 94.24 & 28.03 & 59.53 & 24.63 \\

AirForesight & Seen
& \textbf{56.99} & \textbf{35.83} & \textbf{69.25} & \textbf{30.22}
& \textbf{38.27} & \textbf{38.83} & \textbf{71.26} & \textbf{31.65}
& \textbf{80.08} & \textbf{32.13} & \textbf{66.77} & \textbf{28.45} \\

\midrule

Random & UM
& 202.98 & 0.00 & 0.00 & 0.00
& 158.46 & 0.00 & 0.00 & 0.00
& 265.88 & 0.00 & 0.00 & 0.00 \\

Fixed Action & UM
& 180.47 & 0.52 & 2.61 & 0.39
& 132.89 & 0.89 & 4.28 & 0.67
& 247.72 & 0.00 & 0.25 & 0.00 \\

CMA~\cite{aerialvln} & UM
& 141.68 & 2.30 & 10.02 & 2.16
& 102.29 & 3.57 & 14.26 & 3.33
& 197.35 & 0.50 & 4.03 & 0.50 \\

NavFoM~\cite{navfom} & UM
& 125.10 & 6.30 & 18.95 & 5.68
& 102.41 & 6.77 & 20.07 & 6.04
& 170.58 & 5.36 & 15.71 & 4.97 \\

TravelUAV~\cite{traveluav} & UM
& 106.19 & 9.92 & 36.43 & 8.46
& 77.75 & 11.23 & 40.11 & 9.24
& 146.37 & 8.06 & 31.23 & 7.30 \\

AirForesight & UM
& \textbf{90.30} & \textbf{13.67} & \textbf{41.54} & \textbf{12.02}
& \textbf{67.83} & \textbf{15.86} & \textbf{46.35} & \textbf{13.65}
& \textbf{122.06} & \textbf{10.58} & \textbf{34.76} & \textbf{9.71} \\

\midrule

Random & UO
& 260.14 & 0.16 & 0.16 & 0.16
& 174.10 & 0.48 & 0.48 & 0.48
& 302.96 & 0.00 & 0.00 & 0.00 \\

Fixed Action & UO
& 212.84 & 3.66 & 9.54 & 2.16
& 151.66 & 6.70 & 13.88 & 3.72
& 243.29 & 2.14 & 7.38 & 1.38 \\

CMA~\cite{aerialvln} & UO
& 155.79 & 9.06 & 16.06 & 8.68
& 102.92 & 14.83 & 22.49 & 13.90
& 182.09 & 6.19 & 12.86 & 6.08 \\

NavFoM~\cite{navfom} & UO
& 108.04 & 29.83 & 47.99 & 27.20
& 70.51 & 32.54 & 50.72 & 29.54
& 133.01 & 28.03 & 46.18 & 25.64 \\

TravelUAV~\cite{traveluav} & UO
& \textbf{66.92} & 40.10 & 68.21 & 34.90
& \textbf{44.97} & 42.11 & 71.77 & 35.98
& 77.93 & 39.09 & 66.43 & 34.36 \\

AirForesight & UO
& 66.98 & \textbf{44.36} & \textbf{72.66} & \textbf{39.03}
& 50.11 & \textbf{48.80} & \textbf{75.60} & \textbf{41.57}
& \textbf{75.37} & \textbf{42.14} & \textbf{71.19} & \textbf{37.76} \\

\bottomrule
\end{tabular}%
}
\end{table*}

\subsection{Cross-Space Planning Consistency Loss}
\label{sec:cross_space_consistency}

Future-trajectory prediction provides explicit map-space planning supervision, but it may remain weakly connected to executable navigation if optimized only as a segmentation objective. We therefore introduce a \textbf{cross-space planning consistency loss} (CPCL), which encourages the predicted map-space trajectory to follow the expert navigation direction.

We first extract a local planning direction from the predicted future-trajectory mask $\hat{T}_t$. Because $\hat{T}_t$ covers multiple future steps, we consider only trajectory points within a local neighborhood centered at the UAV position, providing an approximation of the short-horizon navigation intent. Principal Component Analysis (PCA) is applied to these local trajectory points, and the dominant principal axis is selected. Its sign is oriented away from the UAV and normalized to obtain the map-space direction
$\mathbf{v}_{\mathrm{map}}$.
This local estimation is analogous to tangent estimation and is therefore most suitable for locally smooth trajectories.

We derive a stable expert direction from the ground-truth waypoint displacement:
\begin{equation}
\Delta \mathbf{P}_{t}^{\mathrm{gt}}
=
\mathbf{P}_{t+1}^{\mathrm{gt}}-\mathbf{P}_{t}.
\end{equation}
Let $\Pi_{\mathrm{map}}(\cdot)$ denote projection from the 3D action space onto the current UAV-local map plane. The projected expert action direction:
\begin{equation}
\mathbf{v}_{\mathrm{action}}
=
\frac{
\Pi_{\mathrm{map}}
\left(
\Delta \mathbf{P}_{t}^{\mathrm{gt}}
\right)
}{
\left\|
\Pi_{\mathrm{map}}
\left(
\Delta \mathbf{P}_{t}^{\mathrm{gt}}
\right)
\right\|_2
}.
\end{equation}

CPCL minimizes their directional discrepancy using cosine distance:
\begin{equation}
\mathcal{L}_{\mathrm{cons}}
=
1-
\frac{
\mathbf{v}_{\mathrm{map}}^{\top}
\mathbf{v}_{\mathrm{action}}
}{
\left\|\mathbf{v}_{\mathrm{map}}\right\|_2
\left\|\mathbf{v}_{\mathrm{action}}\right\|_2
}.
\end{equation}
Rather than directly aligning the imagined trajectory with the predicted waypoint, this objective uses the expert displacement as a stable directional reference, avoiding error propagation from immature waypoint predictions during early training. Since the waypoint head is supervised by the same expert target, the two branches are guided toward a consistent navigation direction, thereby maintaining the relevance of map-space trajectory imagination to downstream waypoint prediction.

\subsection{Overall Optimization}
\label{sec:overall_optimization}

AirForesight is trained using a two-stage optimization strategy. In the first stage, we optimize only the navigation objective to establish basic waypoint prediction capability:
\begin{equation}
\mathcal{L}_{\mathrm{stage1}}
=
\mathcal{L}_{\mathrm{nav}}.
\end{equation}

In the second stage, waypoint prediction is jointly optimized with current-map reconstruction, future-trajectory prediction, future-map prediction, and cross-space planning consistency:
\begin{equation}
\begin{aligned}
\mathcal{L}_{\mathrm{stage2}}
={}&
\mathcal{L}_{\mathrm{nav}}
+
\lambda_{\mathrm{spatial}}
\left(
\mathcal{L}_{\mathrm{map}}^{\mathrm{cur}}
+
\mathcal{L}_{\mathrm{traj}}
+
\mathcal{L}_{\mathrm{map}}^{\mathrm{fut}}
\right)
\\
&+
\lambda_{\mathrm{cons}}
\mathcal{L}_{\mathrm{cons}}.
\end{aligned}
\end{equation}
Here, $\mathcal{L}_{\mathrm{nav}}$ supervises 3D waypoint prediction;
$\mathcal{L}_{\mathrm{map}}^{\mathrm{cur}}$ and
$\mathcal{L}_{\mathrm{map}}^{\mathrm{fut}}$ supervise current- and future-map prediction;
and $\mathcal{L}_{\mathrm{traj}}$ supervises future-trajectory prediction in map space.
The coefficient $\lambda_{\mathrm{spatial}}$ balances the three spatial-supervision objectives, while $\lambda_{\mathrm{cons}}$ controls the contribution of CPCL.

\section{Experiments}

\subsection{Implementation Details}

\noindent\textbf{Datasets.}
We primarily evaluate AirForesight on OpenUAV~\cite{traveluav}, a target-oriented aerial VLN benchmark with realistic UAV flight dynamics. It contains 12,149 human-piloted trajectories, each paired with five-view RGB observations (front, rear, left, right, and down), waypoint sequences, and expert-refined target descriptions. The benchmark covers 89 object categories and navigation ranges from 50 to 400 meters. Following the standard L1-assistance setting, we evaluate on Test Seen, Test Unseen Map, and Test Unseen Object, each of which contains Full, Easy, and Hard subsets.
To broaden the evaluation beyond OpenUAV, we additionally report results on AerialVLN-S~\cite{aerialvln}, where more recent aerial navigation methods are publicly comparable.

\noindent\textbf{Simulated Environment.}
OpenUAV is built on Unreal Engine 4 and AirSim~\cite{traveluav}, providing photorealistic rendering, UAV dynamics, and multimodal sensor simulation for closed-loop aerial navigation.

\noindent\textbf{Evaluation Metrics.}
Following standard VLN protocols~\cite{metric1,beyondgraph}, we report Navigation Error (NE), Success Rate (SR), Oracle Success Rate (OSR), and Success weighted by Path Length (SPL). 
NE measures the final distance to the target in meters, with lower values being better. SR measures the percentage of successful episodes, while OSR determines whether the agent reaches the success region at any point along its trajectory. SPL further accounts for path efficiency. SR, OSR, and SPL are reported as percentages.

\noindent\textbf{Training Details.}
AirForesight is implemented in PyTorch and trained for 2 epochs on four NVIDIA H100 GPUs with a per-device batch size of 32. We use AdamW~\cite{adamw} with an initial learning rate of $5\times10^{-4}$, cosine decay, and 3\% linear warm-up. Training follows the two-stage strategy described in Sec.~\ref{sec:overall_optimization}: the first stage optimizes only waypoint prediction, while the second jointly optimizes navigation, spatial imagination, and cross-space planning consistency.
Unless otherwise specified, the future horizon is set to $n=10$, the map resolution to $224\times224$, and the number of map query tokens to 9. We set
$\lambda_{\mathrm{spatial}}=0.5$ and
$\lambda_{\mathrm{cons}}=10^{-3}$.

\noindent\textbf{Offline Annotation Pipeline.}
The spatial supervision maps are generated once before training using Gemini 3 Pro, GroundingDINO~\cite{groundingdino}, and MobileSAM~\cite{mobilesam}. Processing all training trajectories requires 7,922 API calls, approximately 7.4 hours, and a total API cost of about \$8.42. These annotation models are not used during model training or inference.
To assess pseudo-label quality, we manually examine 100 sampled trajectories. The resulting category-extraction and spatial-projection accuracies are approximately 94.0\% and 82.5\%, respectively, indicating that the generated maps provide sufficiently accurate supervision for training.

\subsection{Comparison With State-of-the-Art Methods}

\noindent\textbf{OpenUAV.}
We compare AirForesight with Random Action, Fixed Action, CMA~\cite{aerialvln}, NavFoM~\cite{navfom}, and TravelUAV~\cite{traveluav}. Random Action samples waypoints without task-specific planning, while Fixed Action maps instructions to predefined macro-actions. CMA employs recurrent cross-modal attention and a trajectory decoder. NavFoM is a general navigation foundation model, and TravelUAV is an LLM-based aerial navigation framework with hierarchical trajectory generation.

As shown in Table~\ref{tab:final_results}, AirForesight achieves the best results across all metrics and difficulty levels on Test Seen. On the Full subset, it reduces NE from 68.44\,m to 56.99\,m and improves SR from 30.47\% to 35.83\% over TravelUAV. SPL increases from 25.37\% to 30.22\%, corresponding to an improvement of 4.85 percentage points.

AirForesight also shows improved performance under distribution shifts. On Test Unseen Map, it improves all four metrics over TravelUAV, including a 3.75-point SR gain and a 15.89\,m reduction in NE. On Test Unseen Object, it achieves higher SR, OSR, and SPL, while its NE is comparable to TravelUAV (66.98\,m versus 66.92\,m). These results indicate improved generalization relative to existing baselines, although unseen environments remain substantially more challenging than the seen setting.

\noindent\textbf{AerialVLN-S.}
We further compare AirForesight with AerialVLN, STMR, CityNavAgent, and OpenFly on AerialVLN-S. As shown in Table~\ref{tab:aerialvln_s}, AirForesight achieves the highest SR on both validation splits. It also obtains the lowest NE on Validation Seen and remains competitive on Validation Unseen.

\begin{table}[t]
\centering
\caption{Comparison on the AerialVLN-S benchmark.}
\label{tab:aerialvln_s}
\vspace{-2mm}

\renewcommand{\arraystretch}{0.95}
\setlength{\tabcolsep}{4pt}

\resizebox{0.9\columnwidth}{!}{%
\begin{tabular}{lcc|cc}
\toprule
\multirow{2}{*}{Method}
& \multicolumn{2}{c|}{Validation Seen}
& \multicolumn{2}{c}{Validation Unseen} \\
\cmidrule(lr){2-3}
\cmidrule(lr){4-5}
& NE$\downarrow$ & SR$\uparrow$
& NE$\downarrow$ & SR$\uparrow$ \\
\midrule
AerialVLN~\cite{aerialvln}
& 90.2 & 7.2
& 127.9 & 5.1 \\

STMR~\cite{STMR}
& 96.3 & 12.6
& 119.5 & 10.8 \\

CityNavAgent~\cite{citynavagent}
& 80.8 & 13.9
& \textbf{60.2} & 11.7 \\

OpenFly~\cite{openfly}
& 127.2 & 8.1
& 113.8 & 7.6 \\
\midrule

AirForesight
& \textbf{77.8} & \textbf{14.4}
& 86.5 & \textbf{12.4} \\
\bottomrule
\end{tabular}%
}

\vspace{-6mm}
\end{table}

\subsection{Ablation Studies}

Unless otherwise specified, all ablation studies are conducted on the OpenUAV Test Seen (Full) split using the same training configuration as the full model.

\begin{table}[t]
\centering
\caption{Component ablation on OpenUAV Test Seen (Full). CSMM, FSI, and CPCL denote Current Spatial Map Modeling, Future Spatial Imagination, and Cross-Space Planning Consistency Loss, respectively.}
\label{tab:abs_total}
\renewcommand{\arraystretch}{1}
\setlength{\tabcolsep}{5.4pt}
\begin{tabular}{ccc|cccc}
\toprule
CSMM & FSI & CPCL
& NE$\downarrow$
& SR$\uparrow$
& OSR$\uparrow$
& SPL$\uparrow$ \\
\midrule
         &            &            & 74.02 & 31.94 & 63.60 & 26.89 \\
\checkmark &          &            & 62.72 & 35.33 & 65.81 & 29.78 \\
         & \checkmark &            & 63.20 & 33.40 & 65.52 & 28.53 \\
\checkmark & \checkmark &          & 62.28 & 34.71 & 67.07 & 29.36 \\
\checkmark & \checkmark & \checkmark
& \textbf{56.99}
& \textbf{35.83}
& \textbf{69.25}
& \textbf{30.22} \\
\bottomrule
\end{tabular}
\vspace{-3mm}
\end{table}

\subsubsection{Effect of Different Components.}

Table~\ref{tab:abs_total} evaluates the contribution of each component. The baseline in Row 1 follows the same two-stage training strategy but optimizes only waypoint prediction in the second stage, without structured spatial supervision.

Adding \textbf{Current Spatial Map Modeling (CSMM)} improves all metrics, reducing NE from 74.02\,m to 62.72\,m and increasing SR from 31.94\% to 35.33\%. This result supports the usefulness of explicitly supervising a structured representation of the current semantic and geometric layout.

Adding \textbf{Future Spatial Imagination (FSI)}, comprising future trajectory and future map imagination, improves the baseline, demonstrating the value of future-aware spatial supervision for navigation planning. Combining CSMM and FSI further improves NE and OSR relative to either component alone. However, its SR and SPL remain slightly below the CSMM-only setting, suggesting that the independently supervised map-space trajectory and action prediction may still exhibit directional mismatch.

Finally, introducing \textbf{CPCL} produces the best results across all metrics. Relative to CSMM+FSI, CPCL reduces NE from 62.28\,m to 56.99\,m and improves SR, OSR, and SPL from 34.71\%, 67.07\%, and 29.36\% to 35.83\%, 69.25\%, and 30.22\%, respectively. These gains support the benefit of encouraging the map-space trajectory and waypoint branch to follow a shared expert navigation direction.

\begin{table}[t]
\centering
\caption{Effect of the number of learnable map query tokens on OpenUAV Test Seen (Full).}
\label{tab:map_tokens}
\renewcommand{\arraystretch}{1}
\setlength{\tabcolsep}{9pt}
\begin{tabular}{c|cccc}
\toprule
Map Tokens
& NE$\downarrow$
& SR$\uparrow$
& OSR$\uparrow$
& SPL$\uparrow$ \\
\midrule
4  & 76.11 & 33.85 & 61.91 & 28.31 \\
9  & \textbf{56.99} & \textbf{35.83} & \textbf{69.25} & \textbf{30.22} \\
16 & 69.06 & 34.63 & 63.04 & 29.12 \\
\bottomrule
\end{tabular}
\vspace{-4mm}
\end{table}

\subsubsection{Effect of the Number of Map Query Tokens.}

As shown in Table~\ref{tab:map_tokens}, increasing the number of map tokens from 4 to 9 improves all metrics, indicating that additional tokens help encode richer spatial information. Performance declines with 16 tokens, possibly because excessive tokens introduce representation redundancy or increase optimization difficulty. We therefore use 9 map tokens as the default setting.

\subsubsection{Effect of the Spatial-Supervision Weight.}

Table~\ref{tab:loss_weight_ablation} evaluates the sensitivity to $\lambda_{\mathrm{spatial}}$. A small weight of 0.1 provides weaker supervision for learning the structured spatial representations. Increasing it to 0.5 yields the lowest NE and highest OSR while maintaining competitive SR and SPL. A larger weight of 1.0 slightly improves SR and SPL but increases NE. We therefore set $\lambda_{\mathrm{spatial}}=0.5$ as a balanced default.

\begin{table}[t]
\centering
\caption{Sensitivity to the spatial-supervision weight $\lambda_{\mathrm{spatial}}$ on OpenUAV Test Seen (Full).}
\label{tab:loss_weight_ablation}
\renewcommand{\arraystretch}{1}
\setlength{\tabcolsep}{9pt}
\begin{tabular}{c|cccc}
\toprule
$\lambda_{\mathrm{spatial}}$
& NE$\downarrow$
& SR$\uparrow$
& OSR$\uparrow$
& SPL$\uparrow$ \\
\midrule
0.1 & 57.58 & 35.41 & 67.63 & 30.14 \\
0.5 & \textbf{56.99} & 35.83 & \textbf{69.25} & 30.22 \\
1.0 & 59.38 & \textbf{36.11} & 69.18 & \textbf{30.52} \\
\bottomrule
\end{tabular}
\vspace{-4mm}
\end{table}

\begin{figure*}[!th]
    \centering
    \includegraphics[width=\textwidth]{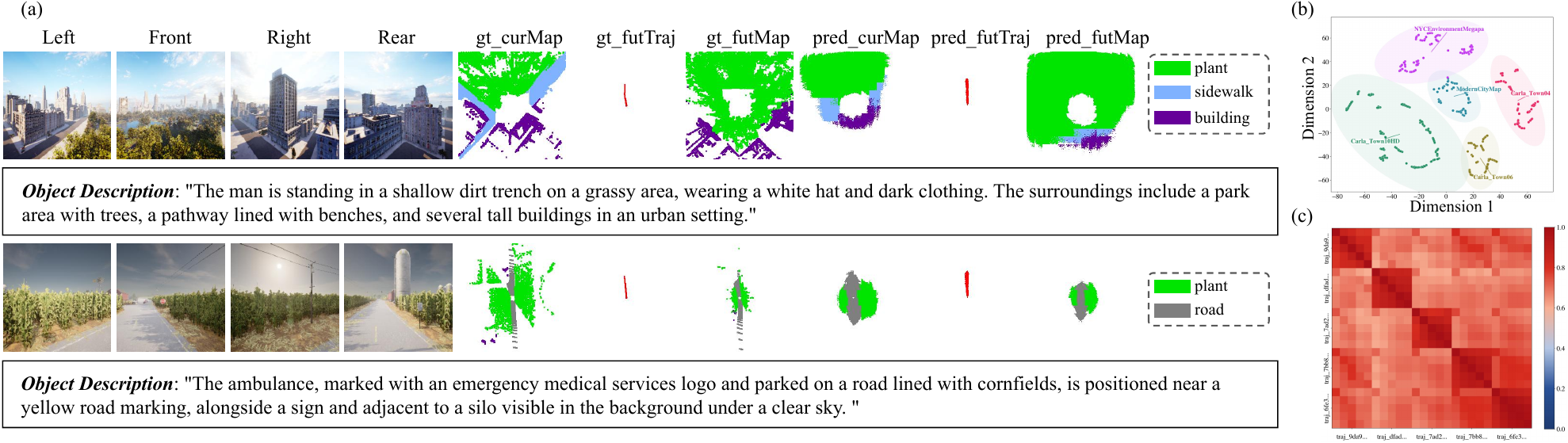}
    \vspace{-6mm}
    \caption{Visualization of learned spatial representations. 
    (a) Predicted current maps, future trajectories, and future maps, together with their corresponding supervision targets. 
    (b) t-SNE projection of map-token representations sampled from different scenes. 
    (c) Cross-correlation analysis of map-token representations from different trajectories within the same scene.}
    \label{fig:map_meaningful}
\end{figure*}

\subsection{Qualitative Analysis}

\begin{figure*}[!t]
    \centering
    \includegraphics[width=\textwidth]{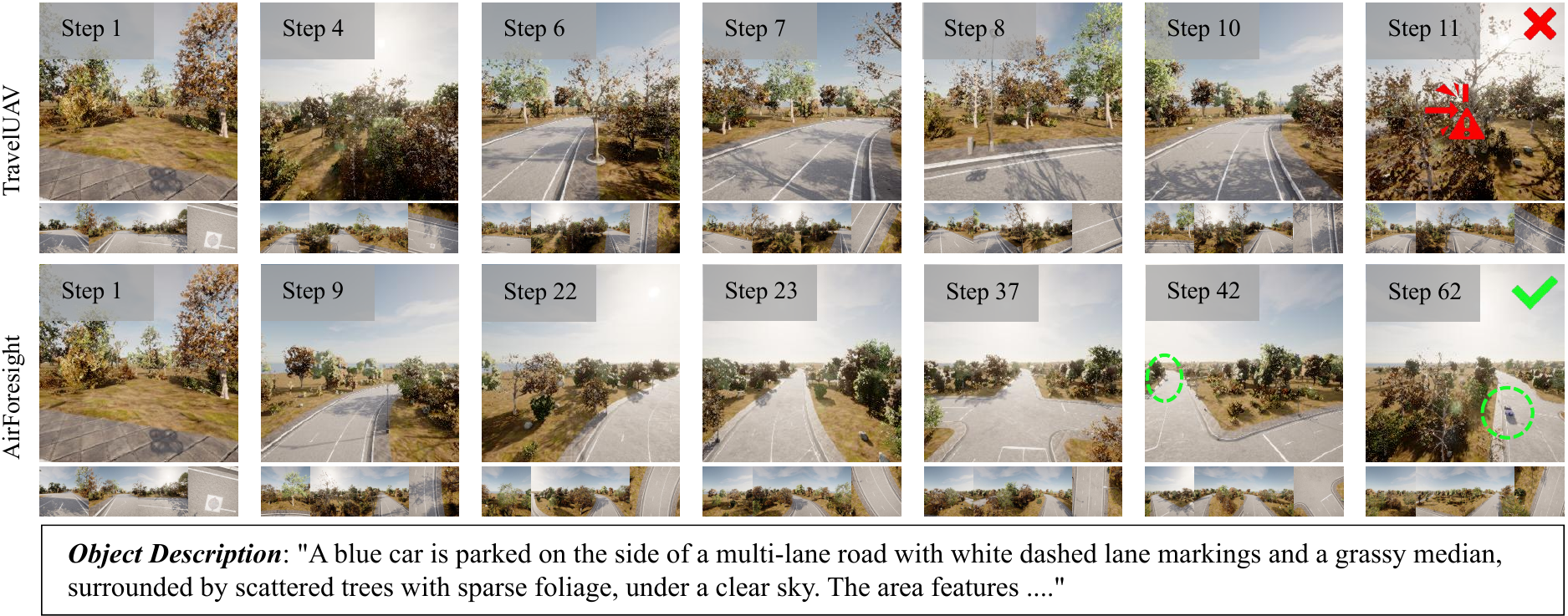}
     \vspace{-4mm}
    \caption{Qualitative comparison on a challenging long-horizon UAV-VLN episode. The top and bottom rows show representative navigation steps of TravelUAV and AirForesight, respectively, under the same instruction.}
    \label{fig:visual_traj}
\end{figure*}

\subsubsection{Qualitative Comparison of Navigation Behaviors}

Figure~\ref{fig:visual_traj} compares TravelUAV and AirForesight on a challenging long-horizon episode. TravelUAV gradually deviates from the feasible route after entering a region with sparse local landmarks and eventually collides with trees. In contrast, AirForesight remains closer to the intended route, passes the intersections shown at Steps 37 and 42, and reaches the target. This example illustrates how structured current-to-future spatial representations can provide useful context for navigation in spatially ambiguous regions. Additional examples are provided in the Appendix.

\subsubsection{Visualization of Latent Map Representations}

To examine the information encoded by the learned map tokens, we visualize their decoded predictions in Figure~\ref{fig:map_meaningful}(a). The predicted current maps capture major semantic categories and their approximate spatial layouts. The future-trajectory and future-map predictions also broadly agree with their supervision targets, providing qualitative evidence that the learned representations encode both scene structure and future motion-related information.

Figure~\ref{fig:map_meaningful}(b) further presents a t-SNE projection of map-token representations sampled from different scenes. Representations from the same scene tend to form local clusters, while samples from different scenes show greater separation, suggesting that the learned tokens retain scene-dependent spatial information.

Figure~\ref{fig:map_meaningful}(c) reports cross-correlation across trajectories from the same scene. Representations sampled within the same trajectory exhibit higher correlation, whereas representations from different trajectories remain positively correlated. This pattern suggests that the learned representations preserve shared scene structure while retaining trajectory-dependent variation.

\begin{figure}[!t]
    \centering
    \includegraphics[width=0.45\textwidth]{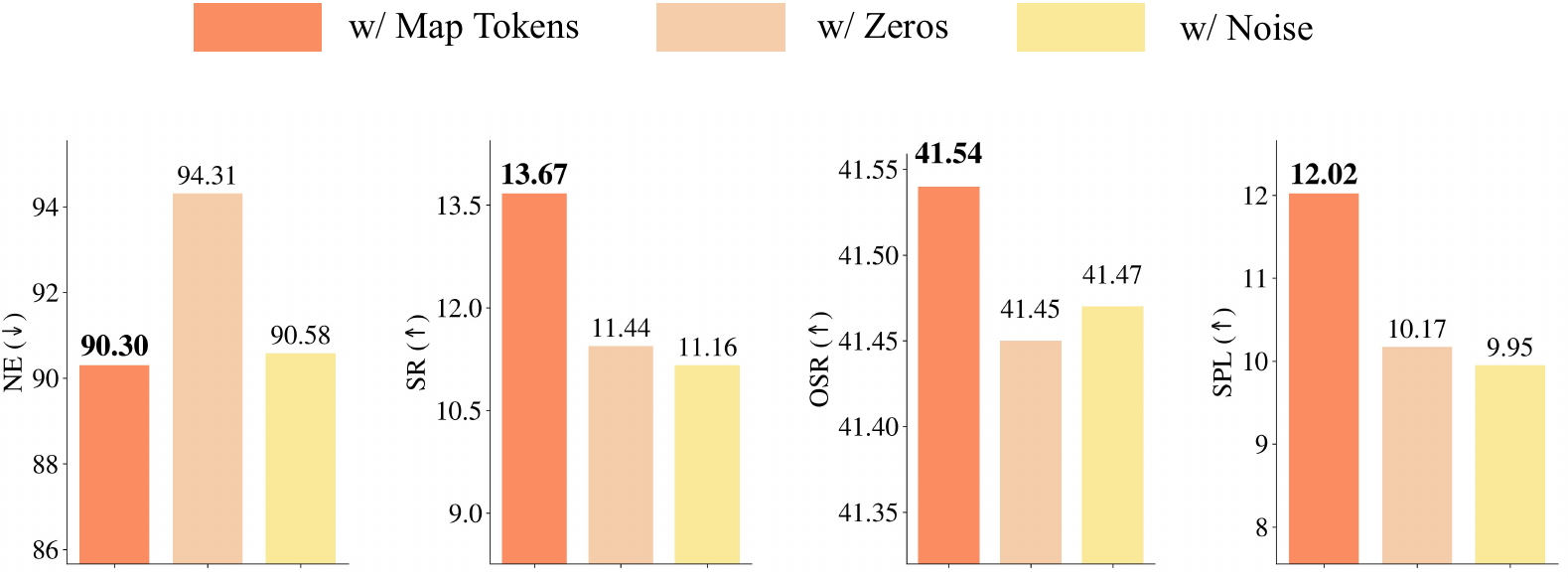}
    \vspace{-2mm}
    \caption{Effect of perturbing the learned map query tokens during inference on OpenUAV Test Unseen Map. The tokens are either retained, replaced with zeros, or replaced with Gaussian noise.}
    \label{fig:map_useful}
    \vspace{-6mm}
\end{figure}

\subsubsection{Use of Latent Map Representations in Navigation}

To examine whether waypoint prediction uses the information carried by the learned map tokens, we perturb these tokens during inference while keeping the remaining model unchanged. Specifically, the map query tokens are replaced with either zero vectors or Gaussian noise on the Test Unseen Map split.
As shown in Figure~\ref{fig:map_useful}, replacing the learned tokens reduces SR from 13.67\% to 11.44\% and 11.16\%, and decreases SPL from 12.02\% to 10.17\% and 9.95\%, respectively. These results suggest that the learned map-token representations contribute to successful and path-efficient navigation, while the model also retains useful navigation cues from the original vision-language inputs.

\section{Conclusion}

We present AirForesight, a current-to-future spatial map imagination framework for UAV-VLN. AirForesight learns latent representations of the current spatial map and future trajectory, progressively infers the future spatial map representations, and integrates these representations for waypoint prediction. We further introduce a cross-space planning consistency loss that encourages agreement between the predicted map-space trajectory direction and the expert action direction derived from the ground-truth waypoint displacement. 
Experiments on OpenUAV and AerialVLN-S, together with detailed ablations, support the effectiveness and stability of the proposed framework. 
Qualitative and perturbation analyses further suggest that the learned map-token representations encode structured spatial information and contribute to downstream navigation.

\begin{acks}
This work was supported in part by the National Natural Science Foundation of China under Grant 62376069, and Grant 92570108, in part by Guangdong Basic and Applied Basic Research Foundation under Grant 2024A1515012027, and in part by Shenzhen Science and Technology Program under Grant KQTD20240729102207002.
\end{acks}

\bibliographystyle{ACM-Reference-Format}
\bibliography{sample-base}

\input{appendix}

\end{document}

%% file: appendix.tex
\appendix

\section{Implementation Details}

\subsection{AirForesight Architecture}

\noindent\textbf{Multimodal Tokenization.}
Given a natural-language instruction $L$ and multi-view RGB observations $I_t^{\mathrm{rgb}}$, the instruction is processed by the language tokenizer, while the visual inputs are encoded by EVA-CLIP~\cite{evaclip} together with a Q-Former. Following LLaMA-VID~\cite{llamavid}, each image is compressed into 17 visual tokens, including one context token for global visual information and 16 content tokens for local details. The language and visual tokens are then concatenated into a unified multimodal sequence.

\noindent\textbf{Learnable Query Tokens.}
AirForesight introduces three groups of learnable query tokens: $\langle\text{cur\_map}\rangle$, $\langle\text{fut\_map}\rangle$, and $\langle\text{waypoint}\rangle$. The current-map and future-map groups each contain nine tokens, while a single waypoint token is used following TravelUAV~\cite{traveluav}. The current-map representation is jointly supervised by current-map reconstruction and future-trajectory prediction, encouraging it to encode both current spatial structure and future motion intent. The future-map tokens model the anticipated future spatial state, while the waypoint token aggregates the preceding multimodal and spatial representations for navigation prediction.

\noindent\textbf{Backbone and Trainable Modules.}
We use Vicuna-7B~\cite{vicuna} as the language-model backbone. Most pre-trained parameters are frozen, while the visual projector, LoRA parameters, spatial decoder, map and trajectory heads, and waypoint decoder are optimized. This parameter-efficient configuration adapts the pre-trained model to UAV-VLN while limiting the number of trainable parameters.

\noindent\textbf{Output Decoders.}
A lightweight MAE-style decoder~\cite{vae} transforms the latent map representations into dense spatial features. 

The latent representations are concatenated with learnable mask tokens equipped with fixed sine--cosine positional embeddings and processed by a shallow stack of Transformer blocks. A shared map-classification head predicts the current and future semantic maps from the corresponding decoded features. 
The future-trajectory mask is predicted by a binary classification head applied to the current-map features, so current-map reconstruction and future-trajectory prediction jointly supervise the same current-map representation. For waypoint prediction, an MLP decoder maps the waypoint-token representation to the normalized 3D direction and distance of the next waypoint.

\subsection{Structured Causal Attention}
\begin{figure}[t]
    \centering
    \includegraphics[width=0.95\columnwidth]{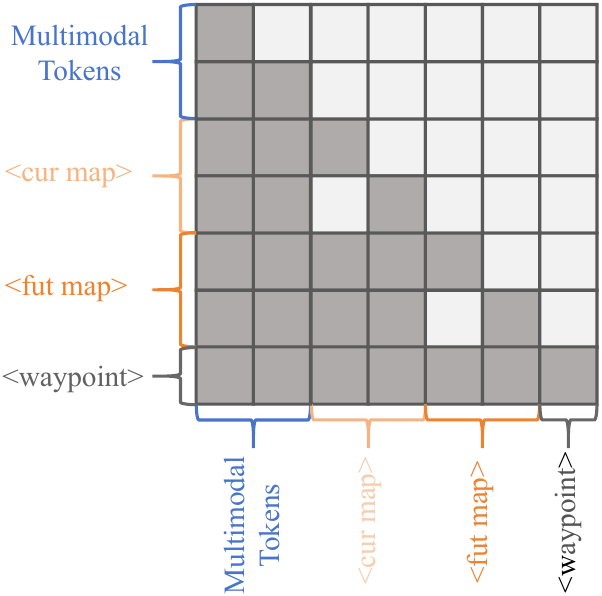}
    \caption{Structured causal attention in AirForesight. Current-map tokens attend to the multimodal inputs; future-map tokens additionally attend to current-map representations; and the waypoint token attends to all preceding representations. Tokens within the same query group are mutually masked.}
    \Description{A causal attention diagram showing multimodal input tokens followed by current-map tokens, future-map tokens, and a waypoint token. Information flows from the multimodal inputs to the current-map tokens, then to the future-map tokens, and finally to the waypoint token.}
    \label{fig:app_causal_mask}
    \vspace{-4mm}
\end{figure}

AirForesight employs structured causal attention to regulate information flow among the multimodal inputs and query-token groups. As shown in Figure~\ref{fig:app_causal_mask}, the current-map tokens attend only to the language and visual inputs. The future-map tokens attend to both the multimodal inputs and the preceding current-map representations, while the waypoint token attends to the multimodal inputs together with the current- and future-map representations.

The future-trajectory mask is decoded from the current-map representation and is not explicitly re-encoded as an input to the future-map or waypoint branches. Instead, future-trajectory supervision encourages the current-map representation to encode future motion intent, which is subsequently propagated through the structured causal attention mechanism.

\begin{table}[!t]
\centering
\caption{Cost and quality of the offline pseudo-label construction pipeline.}
\label{tab:app_annotation_analysis}
\renewcommand{\arraystretch}{1.0}
\setlength{\tabcolsep}{10pt}
\begin{tabular}{lc}
\toprule
Metric & Value \\
\midrule
API calls & 7,922 \\
Cost (\$) & $\sim$8.42 \\
Runtime (h) & $\sim$7.4 \\
Category accuracy (\%) & $\sim$94.0 \\
Spatial-projection accuracy (\%) & $\sim$82.5 \\
\bottomrule
\end{tabular}%
\vspace{-3mm}
\end{table}

\begin{table}[t]
\centering
\caption{Results over three random seeds on OpenUAV Test Seen (Full). $\mu$ and $\sigma$ denote the mean and standard deviation.}
\label{tab:app_multi_seed}
\renewcommand{\arraystretch}{1.0}
\setlength{\tabcolsep}{5pt}
\begin{tabular}{ccccc}
\toprule
Run & NE$\downarrow$ & SR$\uparrow$ & OSR$\uparrow$ & SPL$\uparrow$ \\
\midrule
1 & 60.57 & 34.61 & 68.44 & 29.56 \\
2 & 56.99 & 35.83 & 69.25 & 30.22 \\
3 & 53.64 & 35.34 & 69.70 & 30.29 \\
\midrule
$\mu$ & 57.07 & 35.26 & 69.13 & 30.02 \\
$\sigma$ & 3.47 & 0.61 & 0.64 & 0.40 \\
\bottomrule
\end{tabular}%
\vspace{-4mm}
\end{table}

Within each map-token group, mutual masking prevents tokens of the same type from directly attending to one another. This design encourages individual tokens to extract complementary information from the preceding context rather than relying on within-group interactions.

\subsection{Offline Pseudo-Label Construction}

The semantic maps used for spatial supervision are generated through a one-time offline annotation pipeline. An LLM module extracts task-relevant semantic categories from the navigation instructions and consolidates semantically overlapping labels into a unified set of superclasses. Gemini 3 Pro is used by default, while GPT-4o and Qwen-VL produce comparable category-extraction results on a sampled subset.

The extracted categories are used as prompts for GroundingDINO~\cite{groundingdino}, and categories without confident visual grounding are removed. MobileSAM~\cite{mobilesam} then generates instance masks for the retained objects. Using the corresponding depth observations and camera parameters, the masks are unprojected into a semantic 3D point cloud, transformed into the current UAV-local coordinate frame, and orthogonally projected onto a fixed-scale 2D grid. If multiple semantic points occupy the same grid cell, the label of the highest point is retained.

The same procedure is used to construct the current and future supervision maps. Both $M_t$ and $M_{t+n}$ are represented in the current UAV-local coordinate frame, allowing them to share a consistent spatial reference. The annotation models are used only during offline pseudo-label generation and are not required during model training or inference.

As summarized in Table~\ref{tab:app_annotation_analysis}, processing all training trajectories requires 7,922 API calls, approximately 7.4 hours, and a total API cost of about \$8.42. Manual evaluation of 100 sampled trajectories yields approximately 94.0\% category-extraction accuracy and 82.5\% spatial-projection accuracy. These results demonstrate that the generated pseudo-labels serve as effective spatial supervision signals, despite being imperfect.





\section{Additional Quantitative Analysis}

Unless otherwise specified, the following analyses are conducted on OpenUAV Test Seen (Full).

\subsection{Multi-Seed Evaluation}

Table~\ref{tab:app_multi_seed} reports results from three random seeds. The standard deviations of SR, OSR, and SPL are below one percentage point, indicating that the main navigation results are stable across the evaluated runs.

\subsection{FSI Component Analysis}

\begin{table}[t]
\centering
\caption{Decomposition of Future Spatial Imagination (FSI) on OpenUAV Test Seen (Full).}
\label{tab:app_fsi_decomposition}
\renewcommand{\arraystretch}{1.0}
\setlength{\tabcolsep}{4pt}
\resizebox{\columnwidth}{!}{%
\begin{tabular}{cc|cccc}
\toprule
Future Trajectory & Future Map
& NE$\downarrow$ & SR$\uparrow$ & OSR$\uparrow$ & SPL$\uparrow$ \\
\midrule
\checkmark &            & 67.10 & 33.06 & 64.57 & 28.44 \\
           & \checkmark & 69.32 & 32.62 & 63.90 & 27.81 \\
\checkmark & \checkmark
& \textbf{63.20} & \textbf{33.40} & \textbf{65.52} & \textbf{28.53} \\
\bottomrule
\end{tabular}%
}
\vspace{-2mm}\end{table}

As shown in Table~\ref{tab:app_fsi_decomposition}, future-trajectory and future-map prediction provide complementary supervision. Their combination achieves the best result among the three configurations, supporting the joint modeling of future motion intent and future spatial state.

The benefit of FSI becomes more evident on the Hard split. Adding FSI reduces NE from 93.45\,m to 84.74\,m and improves SR from 28.54\% to 31.81\%, OSR from 60.32\% to 64.41\%, and SPL from 24.98\% to 28.13\%.

\subsection{Effect of Spatial Imagination Form}

\begin{table}[t]
\centering
\caption{Effect of predictive modeling and spatial representation form on OpenUAV Test Seen (Full). ``Pred.'' indicates whether a future state is predicted.}
\label{tab:app_imagination_ablation}
\renewcommand{\arraystretch}{1.0}
\setlength{\tabcolsep}{3pt}
\resizebox{\columnwidth}{!}{%
\begin{tabular}{lcc|cccc}
\toprule
Method & Pred. & Form
& NE$\downarrow$ & SR$\uparrow$ & OSR$\uparrow$ & SPL$\uparrow$ \\
\midrule
Online Spatial Prior
& $\times$ & Map
& 99.51 & 26.52 & 45.98 & 22.40 \\

Local Egocentric Prediction
& \checkmark & Ego.
& 58.63 & 33.82 & 67.92 & 28.48 \\
\midrule

AirForesight
& \checkmark & Map
& \textbf{56.99} & \textbf{35.83} & \textbf{69.25} & \textbf{30.22} \\
\bottomrule
\end{tabular}%
}
\vspace{-4mm}\end{table}

Table~\ref{tab:app_imagination_ablation} compares an online spatial-prior baseline, local egocentric future prediction, and AirForesight. Both predictive variants outperform the non-predictive spatial-prior baseline, supporting the usefulness of future-oriented supervision. AirForesight further outperforms local egocentric prediction across all metrics, suggesting that map-space future representations provide more useful spatial context in this aerial navigation setting.

\begin{table*}[t]
\centering
\caption{Sensitivity to the future horizon, map resolution, and CPCL weight on OpenUAV Test Seen (Full).}
\label{tab:app_additional_sensitivity}
\renewcommand{\arraystretch}{1.0}
\setlength{\tabcolsep}{4pt}
\resizebox{\textwidth}{!}{%
\begin{tabular}{c|cccc|c|cccc|c|cccc}
\toprule
\multicolumn{5}{c|}{Future Horizon}
& \multicolumn{5}{c|}{Map Resolution}
& \multicolumn{5}{c}{CPCL Weight} \\
\midrule
$n$
& NE$\downarrow$ & SR$\uparrow$ & OSR$\uparrow$ & SPL$\uparrow$
& $H\times W$
& NE$\downarrow$ & SR$\uparrow$ & OSR$\uparrow$ & SPL$\uparrow$
& $\lambda_{\mathrm{cons}}$
& NE$\downarrow$ & SR$\uparrow$ & OSR$\uparrow$ & SPL$\uparrow$ \\
\midrule
5
& 67.01 & 32.02 & 60.73 & 26.73
& $112\times112$
& 70.86 & 32.61 & 64.22 & 27.36
& $10^{-2}$
& 60.64 & 35.34 & 67.70 & \textbf{30.29} \\

10
& \textbf{56.99} & \textbf{35.83} & \textbf{69.25} & \textbf{30.22}
& $224\times224$
& \textbf{56.99} & \textbf{35.83} & \textbf{69.25} & \textbf{30.22}
& $10^{-3}$
& \textbf{56.99} & \textbf{35.83} & \textbf{69.25} & 30.22 \\

15
& 60.21 & 33.92 & 67.70 & 28.89
& $336\times336$
& 66.28 & 31.84 & 65.02 & 28.72
& $10^{-4}$
& 58.77 & 34.75 & 65.18 & 29.06 \\
\bottomrule
\end{tabular}%
}
\end{table*}

\subsection{CPCL Alignment Analysis}

To directly assess CPCL, we measure the cosine similarity between the predicted map-space trajectory direction and the expert action direction derived from the ground-truth waypoint displacement. Without CPCL, the similarity is 0.903; adding CPCL increases it to 0.960. This result indicates that CPCL reduces directional mismatch between map-space trajectory prediction and the expert navigation target.

\subsection{Hyperparameter Sensitivity}

Table~\ref{tab:app_additional_sensitivity} shows that an intermediate future horizon of $n=10$ provides the best overall performance. A shorter horizon provides less future context, while a longer horizon increases prediction difficulty. Similarly, the $224\times224$ map resolution achieves the strongest overall result among the evaluated settings. For CPCL, $\lambda_{\mathrm{cons}}=10^{-3}$ yields the lowest NE and the highest SR and OSR, while $10^{-2}$ produces a slightly higher SPL. We therefore use $10^{-3}$ as the balanced default.

\subsection{Efficiency Analysis}

\begin{table}[t]
\centering
\caption{Average inference latency on a single NVIDIA L40S GPU over five independent runs.}
\label{tab:app_inference_efficiency}
\renewcommand{\arraystretch}{1.0}
\setlength{\tabcolsep}{10pt}
\begin{tabular}{lc}
\toprule
Configuration & Latency (ms)$\downarrow$ \\
\midrule
Online Spatial Prior & 1076.39 \\
TravelUAV~\cite{traveluav} & 199.20 \\
AirForesight & 211.51 \\
\bottomrule
\end{tabular}%
\vspace{-4mm}\end{table}

As shown in Table~\ref{tab:app_inference_efficiency}, AirForesight requires 211.51\,ms per inference step, substantially less than the 1076.39\,ms required by online construction and encoding of explicit spatial priors. Relative to TravelUAV, AirForesight introduces 12.31\,ms, or approximately 6.18\%, additional latency. 

\subsection{Effect of Spatial Decoder Architecture}

Table~\ref{tab:app_decoder_ablation} shows that a two-layer decoder with a hidden dimension of 768 achieves the strongest performance among the evaluated settings. A shallower decoder provides less representation capacity, while increasing its depth or width does not improve performance and may introduce additional optimization difficulty. These results motivate the compact spatial decoder used in the final model.

\section{Additional Qualitative Results}

\begin{figure*}[t]
    \centering
    \includegraphics[width=0.95\textwidth,page=1]{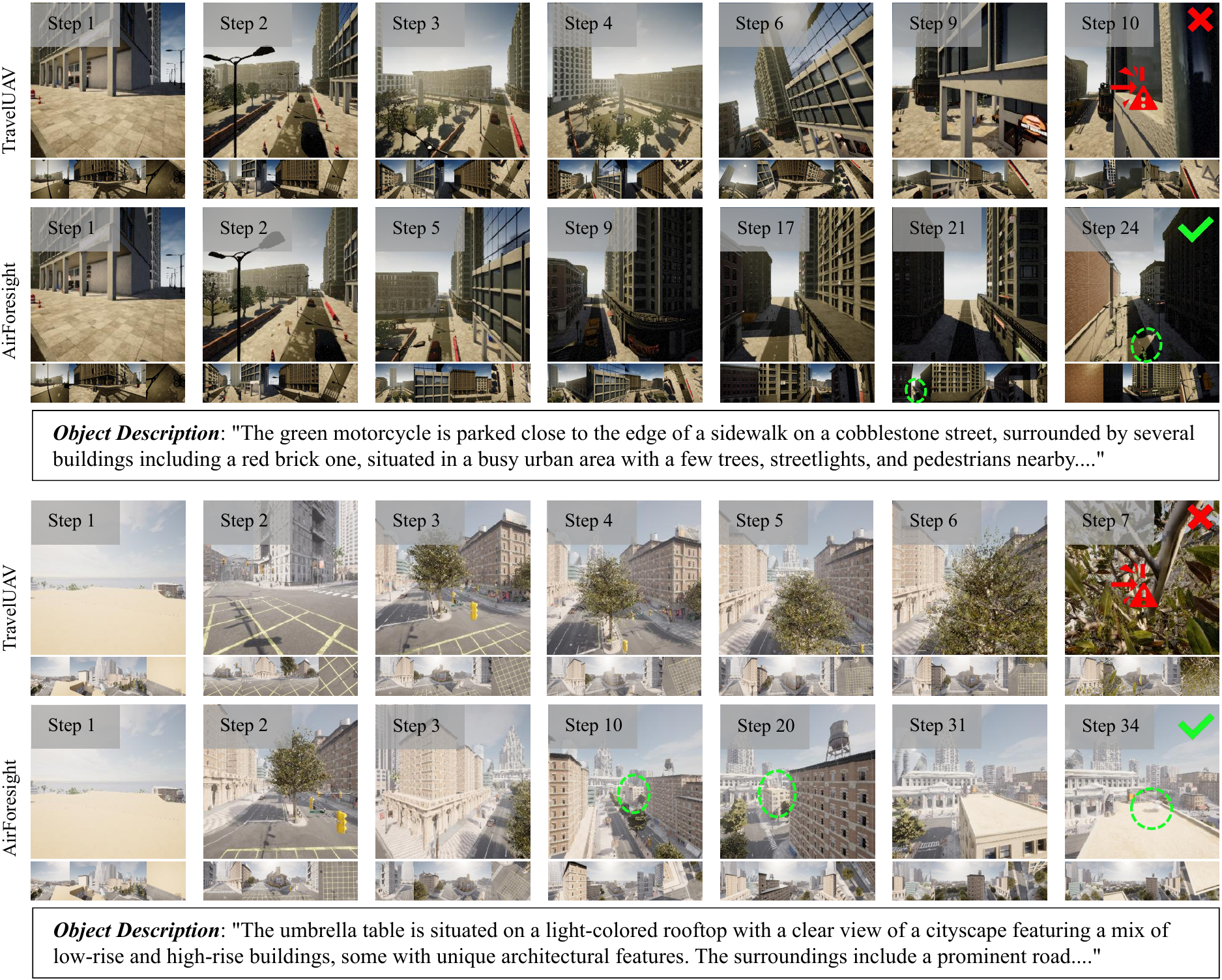}
    \includegraphics[width=0.95\textwidth,page=2]{images/appendix_visual_traj.pdf}
    \caption{Additional qualitative comparisons across diverse long-horizon UAV-VLN episodes. For each example, the top row shows TravelUAV and the bottom row shows AirForesight under the same instruction.}
    \Description{Several pairs of long-horizon navigation examples. Each pair compares representative navigation steps generated by TravelUAV in the top row and AirForesight in the bottom row under the same instruction.}
    \label{fig:app_visual_traj}
\end{figure*}

\subsection{Additional Navigation Examples}

Figure~\ref{fig:app_visual_traj} presents additional navigation examples in cluttered and long-horizon scenes. In the illustrated episodes, AirForesight generally maintains more consistent progress toward the target, while TravelUAV is more likely to deviate in regions with sparse or ambiguous landmarks. These examples complement the quantitative results but should be interpreted as qualitative evidence rather than a complete evaluation of navigation behavior.

\begin{figure*}[t]
    \centering
    \includegraphics[width=\textwidth]{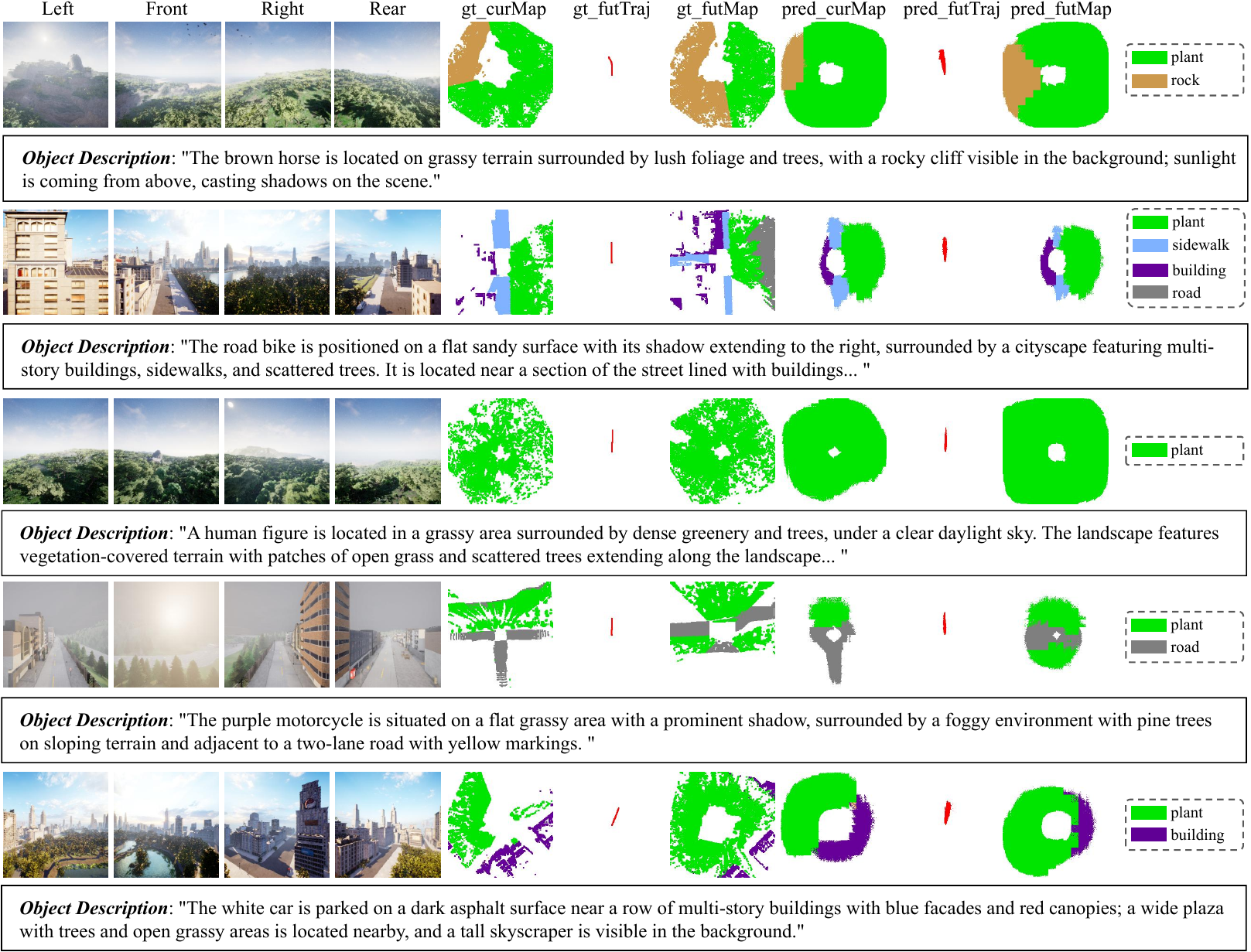}
    \caption{Additional current-map, future-trajectory, and future-map predictions with their corresponding supervision targets.}
    \Description{Examples of multi-view observations followed by predicted semantic current maps, future trajectory masks, and future maps, shown alongside their corresponding supervision targets.}
    \label{fig:app_semantic_map}
\end{figure*}

\subsection{Additional Spatial-Prediction Examples}

Figure~\ref{fig:app_semantic_map} presents additional decoded current maps, future trajectories, and future maps. The current-map predictions capture major semantic categories and approximate spatial layouts, while the future-trajectory and future-map predictions broadly follow their supervision targets. These examples provide qualitative evidence that the map-token representations encode scene-dependent spatial structure and future motion-related information.

Since the ground-truth semantic maps are generated through an automatic annotation process, the generated maps and model predictions may not exactly match the real environment in every detail. Although these maps do not constitute pixel-perfect reconstructions, such discrepancies do not prevent the model from learning meaningful semantic layouts and spatial relationships for planning.

\begin{table}[t]
\centering
\caption{Sensitivity to the depth and hidden dimension of the spatial decoder on OpenUAV Test Seen (Full).}
\label{tab:app_decoder_ablation}
\renewcommand{\arraystretch}{1.0}
\setlength{\tabcolsep}{5pt}
\begin{tabular}{c|cccc}
\toprule
Depth
& NE$\downarrow$ & SR$\uparrow$ & OSR$\uparrow$ & SPL$\uparrow$ \\
\midrule
1 & 61.18 & 35.55 & 67.21 & 30.16 \\
2 & \textbf{56.99} & \textbf{35.83} & \textbf{69.25} & \textbf{30.22} \\
4 & 69.37 & 34.63 & 62.40 & 29.05 \\
\midrule
Hidden Dim.
& NE$\downarrow$ & SR$\uparrow$ & OSR$\uparrow$ & SPL$\uparrow$ \\
\midrule
512  & 68.52 & 33.24 & 62.70 & 27.71 \\
768  & \textbf{56.99} & \textbf{35.83} & \textbf{69.25} & \textbf{30.22} \\
1024 & 77.79 & 33.22 & 60.14 & 27.73 \\
\bottomrule
\end{tabular}%
\vspace{-5mm}\end{table}

\section{Limitations and Failure Modes}

AirForesight remains subject to several limitations. First, its spatial supervision depends on automatically generated pseudo-labels. Errors in semantic-category extraction, open-vocabulary grounding, segmentation, depth observations, or 3D-to-2D projection may propagate into model training. Although the measured pseudo-label quality is relatively high, the supervision remains imperfect.

Second, CPCL estimates a local trajectory direction using PCA. This approximation is suitable for locally smooth trajectories but may be less reliable for sharply curved paths, sparse trajectory masks, or ambiguous local geometry. More flexible curve-aware alignment could be explored in future work.

Third, future-trajectory information is propagated through the shared current-map representation rather than by explicitly re-encoding the predicted trajectory mask into the future-map branch. Direct trajectory-conditioned future-map imagination may provide more fine-grained route-dependent reasoning and is worth investigating.

Finally, performance still decreases substantially on unseen maps and unseen objects, indicating that large environmental and semantic shifts remain challenging. The navigation policy also continues to use visual-language inputs alongside the learned map representations, rather than depending exclusively on the map tokens.